\def\year{2022}\relax
\documentclass[letterpaper]{article} 
\usepackage{aaai22}  
\usepackage{times}  
\usepackage{helvet}  
\usepackage{courier}  
\usepackage[hyphens]{url}  
\usepackage{graphicx} 
\usepackage{natbib}  
\usepackage{caption} 
\DeclareCaptionStyle{ruled}{labelfont=normalfont,labelsep=colon,strut=off} 
\usepackage{algorithm}
\usepackage{algorithmic}
\usepackage{booktabs}       
\usepackage{xcolor}         
\usepackage{times}
\usepackage{tikz}
\usepackage{hyperref}       
\usepackage{url}            
\usepackage{booktabs}       
\usepackage{amsfonts}       
\usepackage{amsmath}
\usepackage{newfloat}
\usepackage{listings}
\usepackage{xspace}
\usepackage{multirow}
\usepackage{color}
\usepackage{enumitem}
\usepackage{subfigure}

\usepackage[textsize=scriptsize]{todonotes}
\definecolor{blue}{RGB}{0, 93, 170}			
\definecolor{darkgreen}{HTML}{3bb35b}

\newcommand{\smalltt}[1]{{\small \texttt{#1}}}
\newcommand{\sparql}{\textsc{sparql}\xspace}
\newcommand{\ask}{\textsc{ask}\xspace}
\newcommand{\select}{\textsc{select}\xspace}
\newcommand{\countsparql}{\textsc{count}\xspace}
\newcommand{\filter}{\textsc{filter}\xspace}

\newcommand{\sysname}{STaG-QA}

\newcommand{\seqseq}{\textsc{Seq2Seq}\xspace}

\floatstyle{ruled}
\newfloat{listing}{tb}{lst}{}
\floatname{listing}{Listing}
\usepackage{xcolor}

\usepackage[T1]{fontenc}
\usepackage[scaled=0.85]{beramono}
\usepackage{listings}
\title{A Two-Stage Approach towards Generalization in Knowledge Base Question Answering}

\author{
    Srinivas Ravishankar$^*$ 
    June Thai$^+$ 
    Ibrahim Abdelaziz$^*$
    Nandana Mihidukulasooriya$^*$ \\
    Tahira Naseem$^*$
    Pavan Kapanipathi$^*$
    Gaetano Rossiello$^*$
    Achille Fokoue$^*$\\
    $^*$IBM Research\quad$^+$UMass Amherst
}

\begin{document}

\maketitle

\begin{abstract}
Most existing approaches for Knowledge Base Question Answering (KBQA) focus on a specific underlying knowledge base either because of inherent assumptions in the approach, or because evaluating it on a different knowledge base requires non-trivial changes. However, many popular knowledge bases share similarities in their underlying schemas that can be leveraged to facilitate generalization across knowledge bases.
To achieve this, we introduce a KBQA framework based on a  2-stage architecture that explicitly separates semantic parsing from the knowledge base interaction, facilitating  transfer learning across datasets and knowledge graphs. 
We show that pretraining on datasets with a different underlying knowledge base can nevertheless provide significant performance gains and reduce sample complexity. Our approach achieves comparable or state-of-the-art performance for LC-QuAD (DBpedia), WebQSP (Freebase), SimpleQuestions (Wikidata) and MetaQA (Wikimovies-KG). 
\end{abstract}

\section{Introduction}

Knowledge Base Question Answering (KBQA) has gained significant popularity in recent times due to its real-world applications, facilitating access to rich Knowledge Graphs (KGs) without the need for technical query-syntax. Given a natural language question, a KBQA system is required to find an answer based on the facts available in the KG. For example, given the question ``Who is the director of the film Titanic", a KBQA system should retrieve the entity corresponding to ``James Cameron". This would be \smalltt{dbr:James\_Cameron}\footnote{dbr: \url{http://dbpedia.org/resource/}} in DBpedia~\cite{auer2007dbpedia}, \smalltt{wd:Q42574}\footnote{wd: \url{http://www.wikidata.org/entity}} in Wikidata~\cite{vrandevcic2014wikidata}, and \smalltt{fb:m.03\_gd}\footnote{fb: \url{http://rdf.freebase.com/ns/}} in Freebase~\cite{bollacker2008freebase}. 


KBQA has been evaluated on multiple different KGs such as Freebase~\cite{bollacker2008freebase}, Wikidata~\cite{vrandevcic2014wikidata}, DBpedia~\cite{auer2007dbpedia}, and MetaQA~\cite{zhang2018variational}. Most existing heuristic-based KBQA approaches such as NSQA~\cite{kapanipathi2020question}, gAnswer~\cite{zou2014natural}, and QAmp~\cite{vakulenko2019message} are typically tuned for a specific underlying knowledge base making it non-trivial to generalize and adapt it to other knowledge graphs. 
On the other hand, WDAqua~\cite{diefenbach2017wdaqua}, a system with a focus on being generalizable, ignores question syntax, thereby showing reduced performance on datasets with complex multi-hop questions. 

Recently, there has been a surge in end-to-end learning approaches that are not tied to specific KGs or heuristics, and hence can generalize to multiple KGs. GrailQA~\cite{gu2021beyond} in particular categorized different forms of generalization, such as novel relation compositionality and zero-shot generalization. They also demonstrated transfer across QA datasets, but within the same KG. 
On the other hand, GraftNet~\cite{sun2018open} and EmbedKGQA~\cite{embedkgqa} demonstrated their ability to generalize over multiple KGs by demonstrating state-of-the-art performance on MetaQA (Wikimovies) as well as WebQSP (Freebase). The two techniques, however, are highly sensitive to the training data; failing to generalize in terms of relation compositionality within a KG. EmbedKGQA and GraftNet show significant drops (between 23-50\%) in performance on relation compositions that are not seen during training. 
Furthermore, it is unclear how these systems transfer across KGs because of their tight-integration with KG-specific embeddings.

In this work, we present a novel generalizable KBQA approach \sysname~ (Semantic parsing for Transfer and Generalization) that works seamlessly with multiple KGs, and demonstrate transfer even across QA datasets with different underlying KGs. Our approach attempts to separate aspects of KBQA systems that are softly tied to the KG but generalizable, from the parts more strongly tied to a specific KG.  Concretely, our approach has two stages: 1) The first stage is a generative model that predicts a query skeleton, which includes the query pattern, the different \sparql operators in it, as well as partial relations based on label semantics that can be generic to most knowledge graphs. 2) The second stage converts the output of the first stage to a final query that includes entity and relations mapped to a specific KG to retrieve the final answer.

Our contributions are as follows:

\begin{itemize}
    \item A simple \seqseq architecture for KBQA that separates aspects of the output that are generalizable across KGs, from those that are strongly tied to a specific KG. 
    \item To the best of our knowledge, our approach is the first to evaluate on and achieve state-of-the-art or comparable performance on KBQA datasets corresponding to four different knowledge graphs, i.e, LC-QuAD (DBpedia), WebQSP (Freebase), SimpleQuestions (Wikidata) and MetaQA (Wikimovies). 
    \item Our extensive experimental results shows that the proposed architecture: (a) facilitates transfer with significant performance gains in low-resource setting; (b) generalizes significantly better (23-50\%) to unseen relation combinations in comparison to state-of-the-art approaches.
\end{itemize}
\section{Proposed Architecture}

The KBQA task involves finding an answer for a natural language question from a given KG. 
Following the semantic parsing techniques for KBQA~\cite{chen-etal-2021-retrack,kapanipathi2020question,yih2015semantic}, we attempt to solve this task by predicting the correct structured \sparql query that can retrieve the required answer(s) from the KG, i.e, by estimating a probability distribution over possible \sparql{} queries given the natural language question.





\begin{figure*}[]
    \centering
    \includegraphics[width=0.9\textwidth]{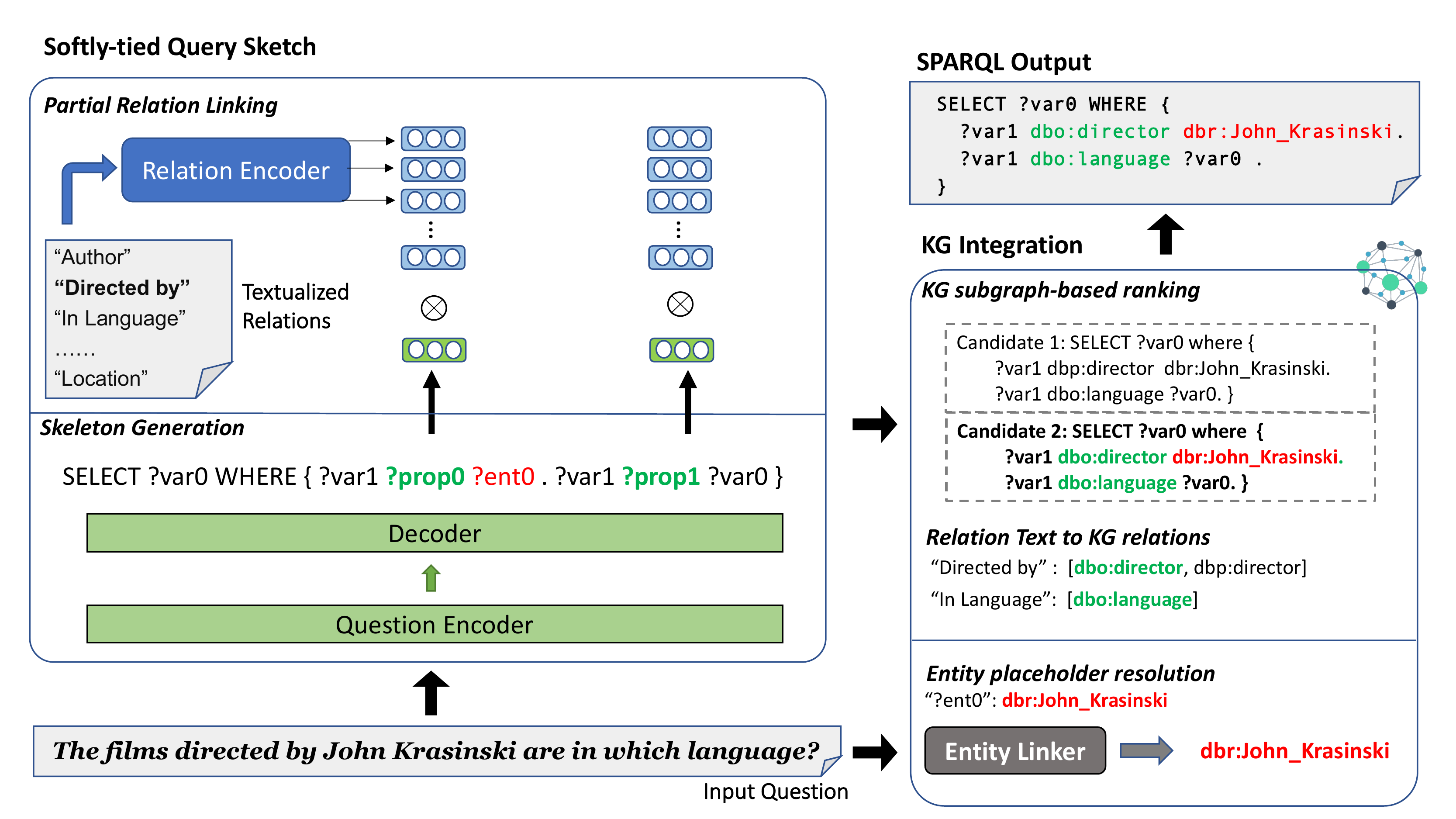}
    \caption{Two-stage system architecture that comprises of: (a) On the left: Softly-tied semantic parse generation that takes an input question return a KG-agnostic parse, and (b) On the right: Knowledge Graph Integration process to eventually return the SPARQL query. }
    \label{fig:arch}
\end{figure*}

In this work, we aim to design a model architecture that generalises across different KGs such as DBpedia, Wikidata, and Freebase. In order to achieve this goal, we have a 2-stage approach as shown in  Figure~\ref{fig:arch}, where we separate generic \sparql query-sketch learning from KG-specific mapping of concepts. Specifically, the following 2-stages are: 

\begin{itemize}
\item \textit{Softly-tied query sketch}: This is the first stage of our approach where we intend to learn aspects of the \sparql query generation that are generic to any knowledge graph. 
Specifically, we observe the following: (i) multi-hop patterns are mostly generic to question answering over KGs.  (ii) across many KGs, analogous relations have semantic or lexical overlap. Therefore, we focus on 2 sub-tasks in this stage, query skeleton generation and partial relation linking. We call the output of this stage a \textit{softly-tied} semantic parse, because the exact output is partially dependent on the specific KG in use, but our choice of representations and architecture ensures that transfer across KGs is a natural consequence. 

\item \textit{KG alignment}: This is the next step where we introduce all vocabulary specific to the knowledge graph in order to generate an executable \sparql query. To do so, we bind the softy-tied semantic parse strongly to the KG to find the answer by (i) resolving the textual relations to KG relations, (ii) introducing KG specific entities into the \sparql skeleton, and (iii) rank the obtained \sparql queries based on its groundings in the KG. 

\end{itemize}


\begin{table}[]
\centering
\begin{tabular}{c|c}
\toprule
KG & Query Graph Structure \\
\midrule 
DBpedia
&
\begin{minipage}{2in}
\small
\begin{verbatim}
?var <director> <entity>
?var <language> ?ans
\end{verbatim}
\end{minipage} \\    
\midrule
Wikimovies
&
\small
\begin{minipage}{2in}
\begin{verbatim}
?var <directed by> <entity>
?var <In language> ?ans 
\end{verbatim}
\end{minipage} \\
\midrule
Wikidata
&
\small
\begin{minipage}{2in}
\begin{verbatim}
?var <director> <entity>
?var <original language 
        of film> ?ans
\end{verbatim}
\end{minipage}
\\
\bottomrule
\end{tabular}
\caption{Query sketch for the question ``The films directed by John Krasinski are in which language?"}
\label{tab:sketch}
\end{table}



\subsection{Softly-tied Query Sketch}

As mentioned above, the goal is to create a representation and architecture that can generalize easily not only across examples within a dataset, but also across KGs. To accomplish this, we define 2 subtasks: (a) Skeleton Generation, and (b) Partial relation linking.



\noindent \textbf{Skeleton Generation:}
A \sparql's  skeleton captures the operators needed to answer the question; i.e. \ask, \select, \countsparql  or \filter,  as well as the query graph structure, with placeholder nodes for entities (e.g. \smalltt{:ent0}), relations (e.g. \smalltt{:prop0}) and variables (e.g. \smalltt{?var0}). For many questions, the generated \sparql skeletons across different KGs are similar, if not identical. The skeleton structures unique to a KG, e.g. reification (present in Wikidata but not DBpedia), can be learnt when fine-tuning on a dataset with that underlying KG. An example of a \sparql skeleton for our running example in Figure~\ref{fig:arch} ``The films directed by John Krasinski are in which language?" is: \\ \\

\begin{lstlisting}[
   basicstyle=\ttfamily,frame=single, xleftmargin=.2\columnwidth, xrightmargin=.2\columnwidth,
   numbers=none]
SELECT ?var0 WHERE { 
    ?var1 :prop0 :ent0 . 
    ?var1 :prop1 ?var0 . 
   }
\end{lstlisting}
\vspace{1.0em}

As shown in Figure~\ref{fig:arch}, the question is passed through a transformer-based \seqseq model which is trained to produce the \sparql skeleton corresponding to the question text. The encoder of the \seqseq model is a bi-direction transformer, while the decoder is auto-regressive with a causal self-attention mask.  

Given a question text, we tokenize it using BERT tokenizer and add special \smalltt{[CLS]} and \smalltt{[SEP]} symbols in the beginning and the end of the question, respectively. This tokenized input is passed through a transformer encoder, producing encoder hidden states for each token at each layer. The encoder is initialized with pretrained BERT model~\cite{devlin2018bert}, which helps generalization with respect to different question syntax.


We then use a transformer decoder with cross attention mechanism. 
At each time step $i$, the decoder considers the encoder states via cross-attention and previous decoder states via self attention. It produces a distribution over possible skeleton output tokens. The decoder output vocabulary $\mathcal{V}$ comprises of entity place holder tokens $\mathcal{V}_e$, relation place holder tokens $\mathcal{V}_r$ and \sparql operators $\mathcal{V}_o$; each of these is a small closed set of tokens.
The output of each decoding step is a softmax over possible operators $s_i \in \mathcal{V}$. Unlike the encoder, no pre-trained model is used for the decoder, and paramerters are initialized randomly.

\textbf{Partial Relation Linking:}
For each relation placeholder in the \sparql skeleton  (\smalltt{:prop0}, \smalltt{:prop1}, etc), we need to identify the appropriate relation that can replace the placeholder to produce the correct semantic representation of the query. 
We have noted previously that relations across KGs share lexical and semantic similarities. For example, in  Table~\ref{tab:sketch} the three KGs (DBpedia, Wikimovies, and Wikidata) represent the relationship ``Directed by" with very similar lexical terms ``Director" and ``Directed by".  We can thus leverage large pre-trained language models to allow generalization and transfer of such relations across KGs. In each KG, we first map the relations to their respective surface forms, using either \smalltt{label} relations from the KG, or by extracting some semantically meaningful surface form from the relation URI. These are the ``textualized relations" shown in Figure \ref{fig:arch}. 
Table \ref{tab:rel2surface} shows some more examples of relation labels for 3 KGs. Note that this mapping can be many-to-one. For example, both \smalltt{dbo:language} and \smalltt{dbp:language} map to the same relation label ``language". 

\begin{table*}
\centering
\begin{tabular}{c|c|c}
\toprule
 KG & KG Relation & Derived Surface Form \\
\midrule
\multirow{3}{*}{DBpedia}    &      dbo:language     &    language \\
            &  dbp:language & language \\
            &  dbp:languages &   languages  \\

\midrule
\multirow{2}{*}{Wikidata}    &  P397  & official language     \\
                             &  P364  & original language of film or TV show  \\
\midrule
\multirow{3}{*}{FreeBase} & people.ethnicity.languages\_spoken & languages spoken\\
& location.country.languages\_spoken &  languages spoken\\
\bottomrule
\end{tabular}
\caption{Examples of textualized relations for different KGs, obtained either using the relation label from the KG (DBpedia, Wikidata) or by extracting a part of the relation URI (Freebase)}
\label{tab:rel2surface}
\end{table*}

 Our goal is to identify which relation surface form best matches each relation placeholder in the skeleton. We thus train the \seqseq decoder and relation encoder to project into the same space. Concretely, the decoder hidden state corresponding to each relation placeholder is optimised to be closest to the encoded representation of the correct relation, using a cross-entropy loss.
 For example, in Figure \ref{fig:arch}, the decoder state for \smalltt{:prop0} should have maximum inner product with the encoded representation for the relation surface form ``Directed by", compared to the encoded representations of all other relations. Our relation encoder is a transformer model whose parameters are initialized with pretrained BERT model. Given that BERT-based representations of lexically or semantically similar relations across KGs will be close, it is easy to see why transfer across KG is possible. The final outcome of partial relation linking is a ranked list of relation surface forms for each placeholder in the skeleton.




The skeleton generation loss and partial relation linking loss are optimized jointly. The \sparql skeleton together with the partial relation linking produces a ranked list of softly-tied query sketches. In the case of multiple placeholders, the score of each pair of relation surface forms is the product of their individual scores.
Sometimes this phase produces multiple semantic interpretations, either due to noisy surface forms (for instance, DBpedia KG includes Wikipedia infobox keys ``as is'' when they can not be mapped to the ontology relations)  or due to the presence of semantically identical or similar relations with distinct identifiers (eg. \smalltt{dbo:language} and \smalltt{dbp:language}).  
For the example, ``The films directed by John Krasinski are in which language?", this stage will produce the following sketches:

\vspace{3mm}
\begin{lstlisting}[frame=single, caption={}, numbers=none,
xleftmargin=.1\columnwidth, xrightmargin=.1\columnwidth]
  P=0.87    SELECT ?var0 where { 
              ?var1 director :ent0. 
              ?var1 language ?var0.}  
  P=0.76    SELECT ?var0 where {
              ?var1 director :ent0. 
              ?var1 languages ?var0.} 
  ...
\end{lstlisting}

\subsection{KG Interaction}

In order to generate an executable \sparql query, we need to introduce vocabulary specific to the KG. The KG interaction stage performs this task. 
Concretely, given a list of candidate query sketches, this stage performs the following steps to produce the final question answer: 1) link the different entities to their corresponding placeholders in the skeleton, 2) disambiguate relations' textual form and link it to the specific KG relations, and 3) select the correct \sparql based on the actual facts in the KG.


In our approach, we leverage a pre-trained off-the-shelf entity linker, BLINK \cite{wu2019zero}. BLINK provides tuples of (surface form, linked entity) pairs. The entity placeholder resolution step aligns the entities with the entity place holders in the query sketch. In the example above, \smalltt{:ent0} will be linked to \smalltt{dbr:John\_Krasinski} in DBpedia, or \smalltt{wd:Q313039} in Wikidata. When multiple entities are present in the question, the position of the corresponding textual span defines the alignment to the entity placeholder variable. During training, the first entity in the question corresponds to \smalltt{:ent0}, the second entity by \smalltt{:ent1}, etc. This pattern is repeated by the system when decoding during inference, making entity placeholder resolution trivial.

The next step is to disambiguate relations' textual form and link them to the specific KG relations. 
Recall from Table \ref{tab:rel2surface} that each surface form in a query sketch can map to one or more KG relations. In our example using DBpedia as a KG, the surface form ``director" could map to both [\smalltt{dbo:director}, \smalltt{dbp:director}] whereas ``language" could map to both [\smalltt{dbo:language}, \smalltt{dbp:language}]. The semantic parsing stage cannot hope to distinguish between these, and thus we rely on the KG to determine the specific relation that should be chosen. Concretely, we replace every relation surface form with each of the possible KG relations it could map to. Thus, each softly-tied query sketch produces one or more fully executable \sparql{}s. For example, the 2 softly-tied sketches from the previous stage in our example produce 4 possible \sparql{}s, see Table \ref{tab:sparql_predictions}. As the final step, we execute the candidate \sparql queries against the KB and choose the highest-ranked \sparql{} that produces an answer for \select queries. Since \ask queries do not necessarily have to be valid in the KG, we only consider the model score in such cases. 



\begin{table}[t]
\centering
\begin{tabular}{c|c}
\toprule
Rank & Predicted SPARQL Query \\
\midrule 
1
&
\begin{minipage}{3in}
\small
\begin{verbatim}
SELECT ?var0 where { 
  ?var1 dbo:director dbr:John_Krasinski.
  ?var1 dbo:language ?var0. }  
\end{verbatim}
\end{minipage} \\    
\midrule
1
&
\small
\begin{minipage}{3in}
\begin{verbatim}
SELECT ?var0 where { 
  ?var1 dbp:director dbr:John_Krasinski. 
  ?var1 dbo:language ?var0. } 
\end{verbatim}
\end{minipage} \\
\midrule
2
&
\small
\begin{minipage}{3in}
\begin{verbatim}
SELECT ?var0 where { 
  ?var1 dbo:director dbr:John_Krasinski. 
  ?var1 dbp:languages ?var0. }  
\end{verbatim}
\end{minipage}
\\
\midrule
2
&
\small
\begin{minipage}{3in}
\begin{verbatim}
SELECT ?var0 where { 
  ?var1 dbp:director dbr:John_Krasinski. 
  ?var1 dbp:languages ?var0. }  
\end{verbatim}
\end{minipage}
\\
\bottomrule
\end{tabular}
\caption{Top predicted SPARQL queries for the question ``The films directed by John Krasinski are in which language?"}
\label{tab:sparql_predictions}
\end{table}

\section{Experiments}

In this section, we compare \sysname~to other state-of-the-art approaches on datasets from multiple KGs. We validate 2 claims: (1) \sysname~achieves state-of-the-art or comparable performance on a variety of datasets and KGs. (2) \sysname~generalizes across KBs and hence facilitating transfer. The results show that pre-training our system  achieves improvement in performance with better gains in low-resource and unseen relation combination settings.

\begin{table}[]
\begin{tabular}{llrrr}
\toprule
               & KG        & Train & Valid & Test \\
\midrule
LC-QuAD 1.0    & DBpedia   &   3,650    &  200     &  1,000    \\
SimpleQuestion & Wikidata  &  15,000     &    2,000   &  2,280   \\
WQSP-FB        & Freebase  &    2898   &    200   &   1,596   \\
MetaQA 1-hop         & Wikimovies &   86,470    &   9,992    &  9,947    \\
MetaQA 2-hop         & Wikimovies &    118,980   &    14,872   &    14,872  \\
MetaQA 3-hop        & Wikimovies &    114,196   &   14,274    &   14,274   \\

\bottomrule
\end{tabular}
\caption{Dataset Statistics}
\label{tab:stats}
\end{table}

\subsection{Datasets}
To evaluate the generality of our approach, we used datasets across a wide variety of KGs including Wikimovies-KG, Freebase \cite{bollacker2008freebase}, DBpedia \cite{auer2007dbpedia}, and Wikidata \cite{vrandevcic2014wikidata}. In particular, we used the following datasets (Table \ref{tab:stats} shows detailed statistics for each dataset): 
(a) \textbf{MetaQA} (Wikimovies-KG) \cite{zhang2018variational} is a large-scale complex-query answering dataset on a KG with 135k triples, 43k entities, and nine relations. It contains more than 400K questions for both single and multi-hop reasoning. 
(b) \textbf{WQSP-FB} (Freebase) \cite{yih2016value} provides a subset of WebQuestions with semantic parses, with 4737 questions in total.
(c) \textbf{LC-QuAD 1.0} (DBpedia) \cite{lcquad}: A dataset with 5,000 questions (4,000 train and 1,000 test) based on templates. It includes simple, multi-hop, as well as aggregation-type questions. LC-QuAD 2.0 is another version of LC-QuAD based on Wikidata. It has 30K question in total and also template-based. Due to the larger underlying KB and the extensive pattern covered, we used LC-QuAD 2.0 dataset for pretraining and showing our transfer results.
(d) \textbf{SimpleQuestions-Wiki} (Wikidata) \cite{simpleq_wd}: a mapping of the popular Freebase's SimpleQuestions dataset to Wikidata KB with 21K answerable questions.

\subsection{Baselines}
In this work, we evaluate against 8 different KBQA systems categorized into unsupervised \cite{kapanipathi2020question, sakor2020falcon, vakulenko2019message, diefenbach2017wdaqua} and supervised approaches \cite{sun2020faithful, maheshwari2019learning, embedkgqa, sun2019pullnet, sun2018open}. 1) NSQA~\cite{kapanipathi2020question,nsqa_demo} is state-of-the-art system for KBQA on DBpedia datasets. 2) QAMP \cite{vakulenko2019message} is an unsupervised message passing approach that provides competitive performance on LC-QuAD 1.0 dataset. (3) WDAqua~\cite{diefenbach2017wdaqua} is another system that generalises well across a variety of knowledge graphs. (4) Falcon 2.0 \cite{sakor2020falcon} is  a heuristics-based approach for joint detection of entities and relations in Wikidata. Since this approach does not predict the query structure, we tested it on SimpleQuestions dataset only. 
(5) EmbedKGQA~\cite{embedkgqa} is the state-of-the-art KBQA system on MetaQA and WebQSP datasets, (6) PullNet~\cite{sun2019pullnet} is recent approach evaluated on MetaQA and WebQSP datasets, (7) GraftNet~\cite{sun2018open} infuses both text and KG into a heterogeneous graph and uses GCN for question answering, and (8) EmQL \cite{sun2020faithful} is a query embedding approach that was successfully integrated into a KBQA system and evaluated on WebQSP and MetaQA datasets.

\subsection{KBQA Evaluation}

Table \ref{tab:results} shows our system results on all four datasets in comparison to existing approaches. 
We show two versions of our system, one pre-trained with LC-QuAD 2.0 datatset \cite{dubey2019lc} (\sysname$_{pre}$) and another trained from scratch on the target dataset only (\sysname). 
%
As noted earlier, to the best of our knowledge, we are the first to show generality across knowledge graphs by evaluating on datasets from DBpedia, Wikidata, Freebase, and Wikimovies-KG. 

Our approach achieves significantly better performance compared to Falcon 2.0 on SimpleQuestions-Wiki dataset with 24\% better F1 score. While Falcon 2.0 is not a KBQA system itself, it jointly predicts entities and relations given a question. Since SimpleQuestions-Wiki requires only a single entity and a single relation, we used Falcon 2.0 output to generate the corresponding \sparql query required for KBQA evaluation. On MetaQA dataset, our system as well as the baselines achieve near perfect scores indicating the simplicity of this dataset. 

On LC-QuAD 1.0, our approach significantly outperforms existing DBpedia-based approaches. When pretrained on LC-QuAD 2.0, the performance is 9\% better F1 compared to NSQA; the state-of-the-art system on DBpedia. 
The large improvement indicates that \sysname\ was able to generalize and learn similar patterns between  LC-QuAD 1.0 and LC-QuAD 2.0.
As for WebQSP, both versions of our approach are inferior compared to EmQL. 
However, it is also worth noting that EMQL also leverages KBC embeddings, which is not currently utilized \sysname\. 
Overall, the results show that \sysname\ shows better or competitive performance on 3 out of  four datasets and when pretrained on another dataset, the performance improves across all datasets. 
In the next section, we analyze different datasets in terms of the degree of challenge they pose for KBQA systems. We propose evaluation splits that will allow us to better discriminate different systems in terms of their performance on these datasets.

\begin{table*}[]
 \centering
 \resizebox{1.7\columnwidth}{!}{%
\begin{tabular}{l|ccc|ccc|c|ccc}
\toprule
\multicolumn{1}{l|}{}             & \multicolumn{3}{c|}{SimpleQuestions-Wiki}                                                                                  & \multicolumn{3}{c|}{LC-QuAD 1.0}                                                                 & \multicolumn{1}{c|}{WebQSP} & \multicolumn{3}{c}{MetaQA}                                                                                                                                            \\ 
\midrule
\multicolumn{1}{l|}{System}       & P                              & R                              & \multicolumn{1}{c|}{F1}                             & P                              & R                              & \multicolumn{1}{c|}{F1}                            & \multicolumn{1}{c|}{Hits@1}  & \begin{tabular}[c]{@{}l@{}}Hits@1\\ 1-Hop\end{tabular} & \begin{tabular}[c]{@{}l@{}}Hits@1\\ 2-Hop\end{tabular} & \begin{tabular}[c]{@{}l@{}}Hits@1\\ 3-Hop\end{tabular} \\
\midrule
\multicolumn{1}{l|}{WDAqua}       & -                              & -                              & \multicolumn{1}{c|}{-}                              & 22.0                           & 38.0                           & 28.0                           & - & - & - & -                                                       \\
\multicolumn{1}{l|}{QAMP}         & -                              & -                              & \multicolumn{1}{c|}{-}                              & 25.0                           & 50.0                           & 33.0       &    -    &    -    &  -& -      \\
\multicolumn{1}{l|}{NSQA}         & -                              & -                              & \multicolumn{1}{c|}{-}                              & 44.8                           & 45.8                           & 44.4  &     -   &     -     &  -  &                 -          \\
\multicolumn{1}{l|}{Falcon 2.0}   & 34                             & 41.1                           & 36.3   & -     & -    & -    &  -      &  -                            &          -      &       -                                                \\
\midrule
\multicolumn{1}{l|}{GraftNet}                          & -          & -          & -                               & -          & -          & -          & 70.3   & 97.0                                                  & 99.9                                                  & 91.4                                                  \\
\multicolumn{1}{l|}{PullNet}                           & -          & -          & -                               & -          & -          & -          & 69.7   & 97.0                                                  & 99.9                                                  & 91.4                                                  \\
\multicolumn{1}{l|}{EmbedKGQA}                         & -          & -          & -                               & -          & -          & -          & 66.6   & 97.5                                                  & 98.8                                                  & 94.8                                                  \\
\multicolumn{1}{l|}{EMQL}                              & -          & -          & -                               & -          & -          & -          & \textbf{75.5}   & -                                                     & 98.6                                                  & 99.1                                                  \\ 
\midrule
\multicolumn{1}{l|}{\sysname}         & 59.4                           & 62.7                           & 61.0                           & \bf{76.5}                           & 52.8                           & 51.4                           & 65.9   & 100.0                                                 & 100.0                                                 & 99.9                                                  \\
\multicolumn{1}{l|}{\sysname$_{pre}$} & \textbf{60.2} & \textbf{63.2} & \textbf{61.7} & {74.5} & \textbf{54.8} & \textbf{53.6} & 68.5   & \textbf{100.0}                                                 & \textbf{100.0}                                                 & \textbf{100.0}                                                 \\ 
\bottomrule
\end{tabular}
}
\caption{Performance against previous state-of-the-art approaches. Following these techniques, we report precision, recall and F1 scores on SimpleQuestions and LC-QuAD 1.0, and Hits@1 performance on WebQSP and MetaQA datasets. The subscript ${pre}$ indicates the ``pre-trained" version of our system using LC-QuAD 2.0 dataset.}
\label{tab:results}

\end{table*}

\subsection{Effect of Pretraining}
\label{sub_sec:pretraining}

\begin{figure}
    \centering
    \includegraphics[width=0.4\textwidth]{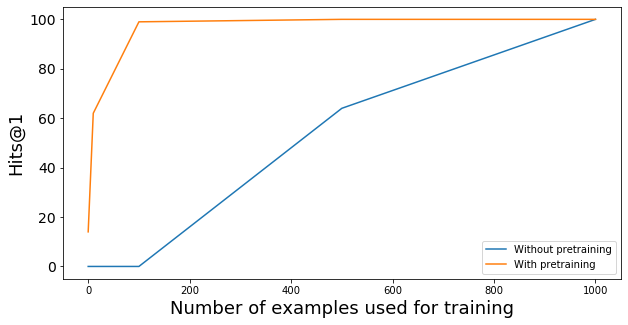}
    \caption{System performance on MetaQA 2-hop questions using different number of training examples, with and without pretraining on LC-QuAD 2.0.}
    \label{fig:pretraining_metaqa}
\end{figure}

\begin{figure}
    \centering
    \includegraphics[width=0.4\textwidth]{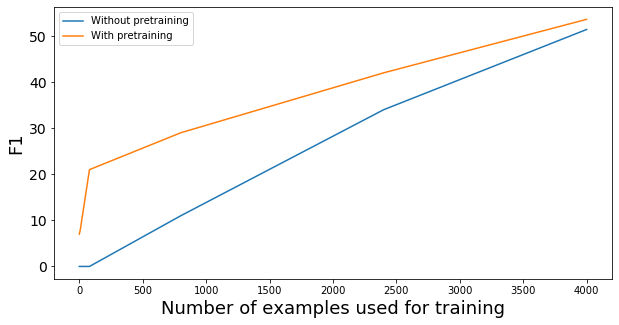}
    \caption{System performance on LC-QuAD 1.0 using different number of training examples, with and without pretraining on LC-QuAD 2.0.}
    \label{fig:pretraining_lcquad1}
\end{figure}


Our architecture is designed to allow transfer learning between entirely different QA dataset/KG pairs. 
As it is harder to show improvements with pre-training on larger datasets, we consider low-resource settings to demonstrate the benefit of transfer, even across KGs. This is useful when there is scarcity of training data for a new target KG. We investigate the benefit of pretraining the semantic parsing stage using LC-QuaD 2.0 (Wikidata KG), before training on the 2-hop dataset in MetaQA (MetaQA-KG) and the LC-QuAD 1.0 dataset (DBpedia). 
Figures \ref{fig:pretraining_metaqa} and \ref{fig:pretraining_lcquad1} show the performance of \sysname\ on each dataset with and without pre-training. We make note of the following  observations. First, without any fine-tuning on either datasets, the pre-trained version  \sysname$_{pre}$\ was able to achieve 18\% Hits@1 on MetaQA and 8\% F1 on LC-QuAD 1.0, indicating the model ability to do zero-shot transfer across knowledge graphs. Second, the pre-trained version provides better performance and is able to converge much faster. For example, in MetaQA (Figure \ref{fig:pretraining_metaqa}), \sysname$_{pre}$ was able to reach almost 100\% Hits@1 with 100 training examples only. To reach the same 100\% Hits@1,  \sysname\ without pretraining required 1,000 examples, an order of magnitude more training data. The same behaviour can be observed on LC-QuAD 1.0, where \sysname$_{pre}$ is better than \sysname, but with both versions continuing to improve as more training data becomes available. 



\subsection{Generalization to novel relation composition}
\label{sub_sec:unseen_relations}

Common KBs have a large number of relations. For example, DBpedia (v2016-10) has around $\sim$60K relations, Wikidata (as of March 2020) has $\sim$8K relations, whereas Freebase contains $\sim$25K relations. In multi-hop queries, these relations can be arranged as paths (e.g., \textit{director} $\rightarrow$ \textit{language}) where possible path combinations grow combinatorially. 
With learning-based approaches, seeing all or most possible relation combinations at training would indeed result in performance improvement at the testing phase. However, this is impractical and hard to enforce in practical scenarios with most KBs as it would require significantly large training data to cover all combinations.  
Instead, an effective KBQA system should be able to generalise to unseen relation paths. In this section, we first analyse existing KBQA datasets to see to which extent this ability is being tested currently. We then create a development set specifically for testing the ability of KBQA systems to generalise to unseen multi-hop relation paths. 

We show in Table~\ref{tab:unseen} the number of test questions in LC-QuAD 1.0, MetaQA and WebQSP datasets that contain relation combinations never seen at training. 
For instance, MetaQA does not test for any unseen relation paths (0\%) where WebQSP contains only 2.06\% of such questions. In contrast, in LC-QuAD 1.0 roughly half of the test questions contain novel relation compositions.

\begin{table}
\centering
\begin{tabular}{l|c} 
\toprule
\textbf{Dataset} & \# of questions with unseen paths   \\ 
\midrule
LC-QuAD 1.0      & 490/1,000 {\small (49 \%)}      \\
WebQSP           & 45/1,638 {\small(2 \%)}    \\
MetaQA 2-hop           & 0/14,872 {\small(0 \%)}\\
MetaQA 3-hop          & 0/14,274 {\small(0 \%)}\\
\bottomrule
\end{tabular}
\caption{Unseen path combinations of seen relations}
\label{tab:unseen}

\end{table}


\paragraph{MetaQA Unseen Challenge Set:} In order to further investigate how this issue affects current KBQA systems, we created a subset from MetaQA, the largest dataset in Table~\ref{tab:unseen} and yet having no unseen relation combinations at testing. We modified the train and dev sets of MetaQA as follows: From the 2-hop training set, we removed training examples containing two randomly chosen relation paths ( \texttt{actor\_to\_movie\_to\_director} and \texttt{director\_to\_movie\_to\_actor}) and split the dev set into two, one containing 13,510 questions with all seen relations path in training and another containing 1,361 questions with all unseen relation paths. 
It is important to note that for each of the unseen relation combinations, the individual relations are present in the training set, i.e, this experiment is designed to test compositionality rather than zero-shot relation linking ability. 

\begin{table}
\centering
\begin{tabular}{c|c|c} 
\toprule
\textbf{System} & 2-Hop Seen & 2-hop Unseen \\
\midrule
EmbedKGQA & 99.00  & 50.00  \\ 
GraftNet-KB & 97.90  & 75.2  \\
GraftNet-Text & 51.2  & 43.3  \\
GraftNet-Both & 99.13  & 95.41  \\
\hline
\sysname & \textbf{99.9} & \textbf{99.7}   \\
\bottomrule
\end{tabular}
\caption{MetaQA Unseen Challenge Set Setting}
\label{tab:unseen_results}
\end{table}

We then trained \sysname, EmbedKGQA and GraftNet on the new reduced training set and tested the performance on our new development sets (seen and unseen). Table \ref{tab:unseen_results} shows the results for each system on 2-hop questions on seen relation paths vs unseen ones. The results clearly demonstrate that there is a significant drop in performance in methods that rank directly across entities in the KG to predict answers. This is most clearly observed in EmbedKGQA, as well as GraftNet-KB, though the use of text (GraftNet-Text and GraftNet-Both) does alleviate this issue. In contrast, our approach is able to maintain exactly the same level of performance for novel relation compositions using KB information alone.

\section{Related Work}
There have been a wide variety of Knowledge Base Question Answering (KBQA) systems~\cite{chen2020formal,embedkgqa,maheshwari2019learning,zou2014natural,diefenbach2017wdaqua,vakulenko2019message,kapanipathi2020question}, trained on datasets that are either question-\sparql pairs (strong supervision) or question-answer pairs (weak supervision). 
More generally, the former can use any logical form that expresses the question as an RDF-query, which is then run on the KG to retrieve the answer. 
We describe some KBQA systems in this section. 

As mentioned above, the first category of KBQA approaches focus on translating the natural language questions into an intermediate logical form to retrieve results from the knowledge base. Generating this kind of semantic parse of the question has shown improved performance compared to weak-supervision based approaches~\cite{yih2016value}. Furthermore, the intermediate structured representation of the question provides a level of interpretability and explanation that is absent in systems that directly rank over entities in the KG to produce answers. This category can further be classified into (a) rule-based approaches such as gAnswer~\cite{hu2017answering}, NSQA~\cite{kapanipathi2020question}, and (b) learning-based approaches such as MaSP~\cite{shen2019multi} and GrailQA~\cite{gu2021beyond}. 

The rule based approaches primarily depend on generic language based syntactic~\cite{zou2014natural} or semantic parses~\cite{nsqa_demo,kapanipathi2020question} of the question and build rules on it to obtain a query graph that represents the \sparql query. NSQA, the state of the art approach for DBpedia based datasets such as LC-QuAD-1.0~\cite{lcquad} and QALD-9~\cite{qald7}, falls in this category. The system uses Abstract Meaning Representation (AMR) parses of the question and a heuristic-based graph-driven methodology to transform the AMR graph to a query graph that represents the \sparql query. 
Many of these systems have components or aspects that are specific to the KG they evaluate on, and do not trivially generalize to other KGs. 
In particular GAnswer, NSQA, and QAmp are specific to DBpedia and do not evaluate their approaches on any other KGs. 

On the other hand, MaSP is a multi-task end-to-end learning approach that focuses of dialog-based KGQA setup. 
MaSP uses a predicate classifier which makes transfer across KGs non-trivial. We adapt the architecture to make it generalizable across KGs, by replacing the relation classifier with a BERT-based ranker that leverages similarities in label semantics between KGs. 
A prominent work, Krantikari~\cite{maheshwari2019learning} is on regular KBQA and has a ranking based approach that is heavily dependent on the knowledge graph. The approach ranks all candidate graph patterns retrieved from the knowledge graph based on the grounded entity. 
In multi-hop settings, as in MetaQA with 3-hop questions, retrieving all possible candidates upto $n$-hops (for an arbitrary choice of $n$) and then ranking across all of them is expensive. In contrast, our work focuses on a generative approach to modeling query graph patterns.
GrailQA is an end-to-end learning approach that generates logical S-forms. It characterizes a few forms of generalization such as compositionality and zero-shot generalization. It also introduces a method that transfers across QA datasets, but within the same KG. On the other hand, we demonstrate transfer across KGs. 

The final category of KBQA approaches are trained solely with question-answer pairs ignoring the intermediate logical representation of the question. EmbedKGQA~\cite{embedkgqa} and GraftNet are two such approaches that directly ranks across entities in the knowledge base to predict an answer, by leveraging either KG embeddings from Knowledge Base Completion (KBC); or creating a unified graph from KB and text. However, these approaches do not generalize well to novel relation compositions not seen during training. Finally, it is unclear how to transfer KBC embedding-based approaches such as EmbedKGQA across KGs since the learnt KG embeddings are tightly coupled with the specific KG in question.

\section{Conclusion}

In this work, we show that a simple 2-stage architecture which explicitly separates the KG-agnostic semantic parsing stage from the KG-specific interaction can generalize across a range of datasets and KGs. We evaluated our approach on four different KG/QA pairs, obtaining state-of-the-art performance on MetaQA, LC-QuAD 1.0, and SimpleQuestions-Wiki; as well as competitive performance on WebQSP. Furthermore, we successfully demonstrate transfer learning across KGs by showing that pre-training the semantic parsing stage on an existing KG/QA-dataset pair can help improve performance in low-resource settings for a new target KG; as well as greatly reduce the number of examples required to achieve state-of-the-art performance. Finally, we show that some popular benchmark datasets do not evaluate generalization to unseen combinations of seen relations (compositionality), an important requirement for a question answering system.
\bibliography{neurips}

\end{document}